\pdfoutput=1

\documentclass[11pt]{article}

\usepackage[]{acl}

\usepackage{times}
\usepackage{latexsym}

\usepackage[T1]{fontenc}

\usepackage[utf8]{inputenc}

\usepackage{microtype}
\usepackage{amsmath,amssymb}
\usepackage{times}
\usepackage{latexsym}
\usepackage{graphicx}
\usepackage{multirow}
\usepackage{booktabs} 
\usepackage{tablefootnote}
\usepackage{colortbl} 
\usepackage{paralist} 
\usepackage{soul} 
\usepackage{subfigure}
\usepackage{xspace}

\DeclareMathOperator*{\argmin}{arg\,min}

\definecolor{lightgrey}{HTML}{dcdbdb}
\sethlcolor{lightgrey}

\usepackage{hyperref}
\usepackage{xspace}
\newcommand{\icl}{ICL\xspace}
\newcommand{\problem}{self-adaptive in-context learning\xspace}
\newcommand{\sicl}{self-adaptive ICL\xspace}

\title{Self-Adaptive In-Context Learning: An Information Compression Perspective for In-Context Example Selection and Ordering}

\author{
Zhiyong Wu$^{\diamondsuit \dagger}$, 
Yaoxiang Wang$^{\clubsuit \dagger}$\thanks{\, Work done while interning at Shanghai AI Lab.}\;, 
Jiacheng Ye$^{\spadesuit}\thanks{\, Equal Contribution.}^{\;\;*}\;$,
Lingpeng Kong$^{\spadesuit}$
\\
$^\diamondsuit$Shanghai Artificial Intelligence Laboratory \quad
$^\clubsuit$Xiamen University \quad
$^\spadesuit$The University of Hong Kong \\
\texttt{\{jcye2,lpk\}@cs.hku.hk, \{wuzhiyong,wangyaoxiang\}@pjlab.org.cn,}
}

\begin{document}
\maketitle

\begin{abstract}
Despite the impressive few-shot performance of in-context learning (ICL), it remains a common practice to randomly select examples to serve as the context. In this paper, we advocate self-adaptive in-context learning, a new principle for ICL, in which the self-adaption mechanism is introduced to help each input find an in-context example organization (i.e., selection and permutation) that can derive the correct output, thus maximizing performance. To validate the effectiveness of self-adaptive ICL, we propose a general select-then-rank framework and a set of novel selection and ranking algorithms. Upon extensive evaluation on eight different NLP datasets, our self-adaptive ICL method achieves a 40\% relative improvement over the common practice setting. Further analysis reveals the great potential of self-adaptive ICL as a promising method to close the gap between ICL and finetuning.  \href{https://github.com/Shark-NLP/self-adaptive-ICL}{\textit{Our code}} will be released to facilitate future research.
\end{abstract}

\section{Introduction}

The increasing scale of pre-trained language models (PLMs) has brought emergent abilities~\citep{wei2022emergent} via in-context learning (\icl), where the PLMs learn to do  downstream tasks simply by conditioning on a prompt containing a few examples of their kinds~\cite{brown2020language}. Due to its impressive performance, \icl has now emerged as a popular and efficient way of  using PLMs. However, \icl is inherently unstable: given different prompts, the performance of \icl on downstream tasks can vary from almost random to comparable with state-of-the-art systems~\cite{zhao2021calibrate,lu2022fantastically,gao2021making}, depending on the quality of the prompts. 

The instability of \icl motivates researchers to explore methods that search for high-performing prompts. 
Note that a \textit{prompt} within the context of \icl contains two ingredients: some input-output pairs (i.e., \textit{in-context examples}) and a \textit{template} that wraps these examples into a natural language instruction. Extensive research has been carried out on searching for a better  template~\cite{gao2021making,shin2020autoprompt,sorensen2022information,deng2022rlprompt}. In contrast, very few efforts have been spent on searching for the best in-context example \textit{organization}. \footnote{In this paper, we abuse the word organization to represent both the selection and ordering of examples.} Recent work, however, has pointed out that the organization of in-context examples can have a significant influence on \icl's performance~\cite{lu2022fantastically,liu2022makes,rubin2022learning}. 

\begin{figure}
    \centering
    \includegraphics[width=\linewidth]{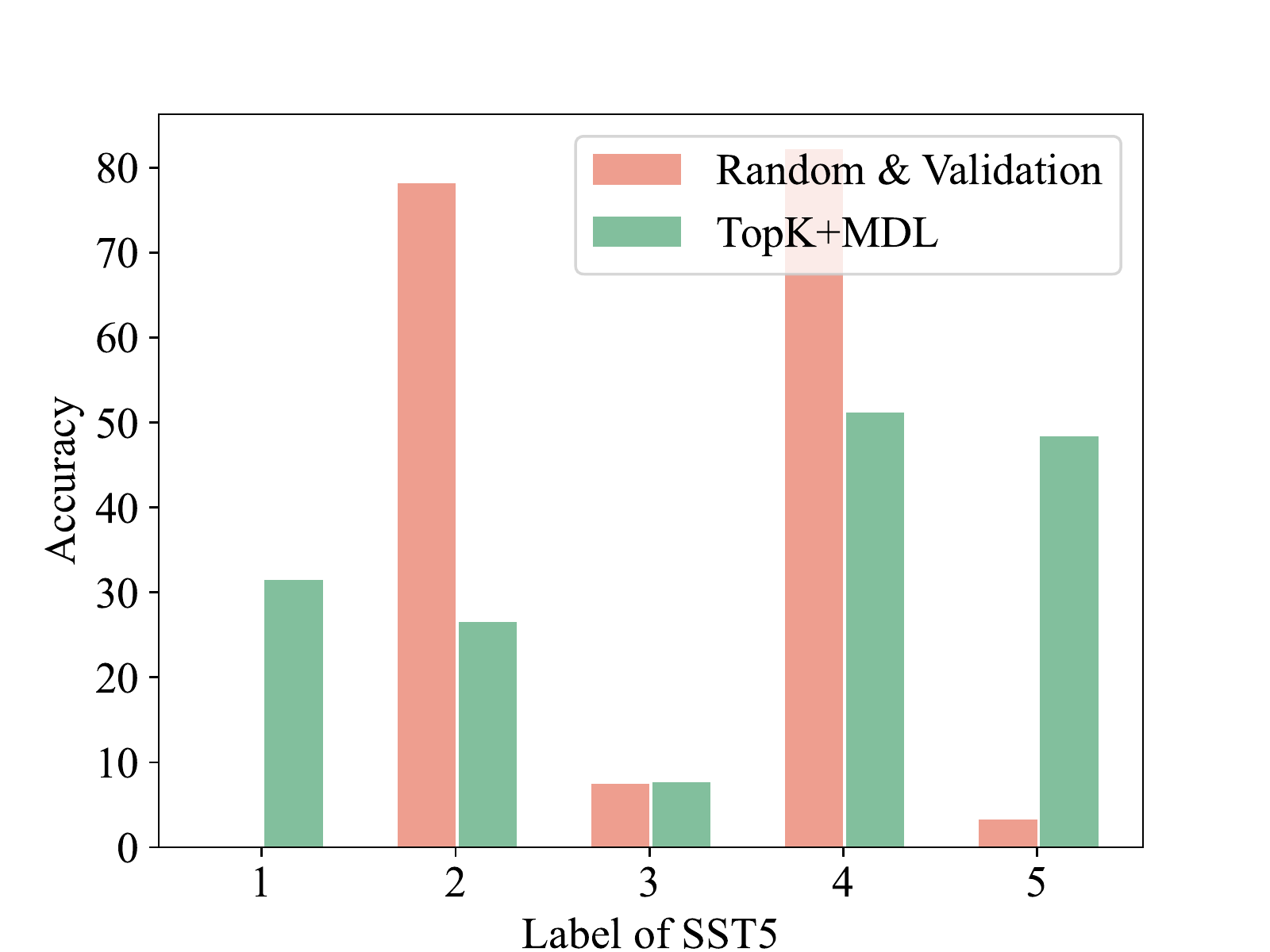}
    \caption{Corpus-level method (red bar) is highly biased towards majority classes, given 8 in-context examples labeled as 2 5 4 4 4 1 2 3.}
    \label{fig:intro}
\end{figure}

This paper fills this gap by proposing a framework for in-context example searching and ranking.
While one can also trivially extend template searching methods to conduct in-context example searching, these methods operate at the \textit{corpus-level}. 
They first construct a small candidate template set using PLMs~\cite{gao2021making,shin2020autoprompt}, data mining algorithms~\cite{jiang2020can}, or by hands~\cite{sorensen2022information}. After that, each candidate will be applied to the whole validation set for inference. According to validation performance, the best template will be adapted for testing. However, existing solutions have the following problems: (i) Their performance relies heavily on the availability of a large-scale high-quality validation set; (ii) Corpus-level methods can be sub-optimal (see Figure~\ref{fig:intro}) because finding a universal template that suits all testing samples perfectly is unlikely. Such majority bias~\cite{zhao2021calibrate} will significantly hurt user experience in practice and make corpus-level methods less robust.

To tackle these issues, we seek to construct a good-performing in-context example organization for each testing sample individually, without access to a validation dataset. This problem, namely \problem, is essentially an NP-hard combinatorial optimization problem that cannot be solved within polynomial time. 
We thus formulate it as a search problem and propose a general two-stage framework to cope with the issue of massive search space.

In the first stage, we apply heuristic rules (e.g., nearest neighbors based on semantic similarity) to filter candidate examples. Given a much smaller candidate set, we then apply algorithms to rank different organizations and look for the best-performing one. Our ranking algorithms are theoretically supported by the Minimal Description Length (MDL) principle and can shed light on why certain permutations are better than others. 




Our contributions are summarized as follows:
\begin{itemize}
    
\item To the best of our knowledge, we are the first to formally define the problem of \problem and formulate it as a two-stage search problem. We propose a general framework to address this problem. 

\item We achieve state-of-the-art performance using the proposed framework and outrun the previous best-performing methods by a large relative improvement. We also find that instance-level ICL methods are generally more robust than corpus-level counterparts. Such empirical success shows a great promise of \sicl. 

\item We conduct extensive analysis for self-adaptive \icl and make some exciting findings. For instance, in Section~\ref{sec:oracle} we reveal that self-adaptive \icl still has much room for improvement. With better search methods, we might be able to close the gap between \icl and finetuning. 

\item We will open-source the proposed framework to facilitate future research. This unified framework enables researchers to identify important design choices in previous methods and paves the way for further improvements. 

\end{itemize}


\section{Related Work}
Despite the surprising zero-shot performance of PLMs, recent works show that \icl can bring the performance to the next level. Augmenting PLMs with \icl achieves SOTA results on a wide range of NLP tasks, ranging from question answering~\cite{joshi2017triviaqa}, information retrieval~\cite{tay2022transformer}, math word problem~\cite{cobbe2021training}, commonsense reasoning~\cite{geva2021did}, and fact checking~\cite{rae2021scaling} etc. The instability of \icl, however, has encouraged researchers to explore methods that search for robust and high-performing prompts. These methods can be categorized as follows based on the target of searching/optimization:

\paragraph{Template search} focuses on searching for the template that can guide PLM's behavior and steer its best performance. Great advances have been made in template searching using various methods: PLMs~\cite{gao2021making}, heuristic rules~\cite{jiang2020can, shin2020autoprompt,prasad2022grips,xu2022zeroprompt}, reinforcement learning~\cite{deng2022rlprompt}, genetic algorithms~\cite{kumar2021reordering}, or by hands~\cite{sorensen2022information,zhao2021calibrate}. 
Nonetheless, all these methods require a high-quality validation set to do prompt selection or optimization.  Unlike them, our framework does not require a validation set. 

When the validation set is not available, researchers propose to search prompts using entropy~\cite{lu2022fantastically} or mutual information~\cite{sorensen2022information}. 
It's worth mentioning that these two works and all aforementioned methods search at the \textit{corpus-level}: they pick the best-performing template with or without a validation set and then equally apply this template to all test examples during inference. However, corpus-level methods might be sub-optimal. If we consider the \textit{No Free Lunch Theorem}, finding one single template that works well for all testing examples is nearly impossible. 

\paragraph{In-context example search,} unlike template search, is rarely explored in the literature despite that they also have a huge impact on \icl performance~\cite{zhao2021calibrate,lu2022fantastically}. \citet{lu2022fantastically} first propose a learning-free corpus-level method for in-context example search. However, they only consider an impractical setting with only 4 examples and their 24 permutations ($^4P_4=4!=24$). \citet{liu2022makes} find examples that are semantically similar to a test sample can serve as a good choice for its in-context examples. However, the reason why such a simple heuristic works is unclear. \citet{su2022selective} extend this nearest neighbor search and further take the diversity of examples into consideration. Inspired by these methods, recent studies propose to learn to retrieve in-context examples~\cite{rubin2022learning}.

\section{Problem formulation}
Given a test sample $(\mathbf{x}, y)$, the probability of generating the target $y$ using a casual PLM $\mathcal{P}$ can be formulated as follows:
\begin{equation}
\label{eq:icl}
p(y|\mathbf{x}) = \mathcal{P}\left(\mathcal{V}(y) | c, \mathcal{T}(\mathbf{x})\right),    
\end{equation}
where $\mathcal{T(\cdot)}$ is the template used to wrap up inputs and $c = \mathcal{T}(\mathbf{x}_1),\cdots, \mathcal{T}(\mathbf{x}_k)$ is the context string concatenating $k$ input-output examples. To deal with classification tasks, a verbalizer $\mathcal{V(\cdot)}$ is introduced to map each label/class $y$ to a word/words in $\mathcal{P}$'s vocabulary. Note that in a special scenario when $k=0$, \icl degenerates to zero-shot \textit{prompting}~\cite{ye2022zerogen,brown2020gpt3}.

The goal of \sicl is then to find an optimal organization of $c \in \mathcal{C}$ that can drive the correct $y$ for each input $\mathbf{x}$, and maximize the task performance. We formulate this as a combinatorial optimization problem.

\section{Method}

In this section, we propose a two-stage framework to tackle the problem of \sicl. 

\subsection{Overview}
In such a combinatorial optimization problem, an exhaustive search is not tractable. So we need specialized algorithms that can quickly rule out large parts of the search space. We present an overview of our selection-then-rank framework here:  
We first use a selection module to reduce the search space. One straightforward choice for pre-ranking would be to use nearest-neighbor algorithms to select examples that are semantically similar to test samples. 
The results are then fed into the ranking module, which picks the best combination and permutation according to information-theoretic-driven criteria. 

\subsection{Selection}
The goal of selection module is to filter out large parts of ``less useful'' examples and construct a small candidate set to reduce the search space. We present various selection methods below.

\paragraph{TopK} \citet{liu2022makes} and \citet{gao2021making} observe that context examples that are closer to the test sample in the embedding space consistently give rise to stronger performance. This observation leads to the TopK method which uses the nearest neighbors of a given test sample as the corresponding in-context examples. 

\paragraph{VoteK} Although \icl was originally proposed for few-shot settings, they often require a large example set to achieve good performance. VoteK~\cite{su2022selective} proposes to alleviate this problem by selecting diverse yet representative examples. Intuitively, VoteK is built upon TopK, but it increases diversity by penalizing examples similar to those already selected. 

\paragraph{DPP} Inspired by VoteK, we also experimented with the determinantal point process (DPP) based method, which is proposed for set selection problems where diversity is preferred. We refer readers to \citet{kulesza2011dpp} for details of DPP.

\subsection{Ranking}
\label{sec:ranking}
With the candidates returned by the selection module, the goal of the ranking module is to determine the best organization among candidates. Our ranking algorithm is inspired by the compression viewpoint of Solomonoff’s general theory of inference~\cite{solomonoff1964formal} and Minimum Description Length (MDL) principle~\cite{grunwald2007minimum} from information theory. 

Both Solomonoff’s theory and the MDL formalize Occam’s razor and hold that a good model of data is a model that is good at losslessly compressing the data, including the cost of describing the model itself. These theories have led to advances in VAE~\cite{kingma2013vae}, and information bottleneck methods~\cite{tishby2015ib}. Inspired by the compression viewpoint of learning, we recast the problem of \problem into a similar paradigm. We assume that a good organization of in-context examples is the  organization that is good at losslessly compressing testing samples. This allows us to give a clear optimization objective when searching for the best organization $c^*$: 
\begin{equation}
    c^* = \argmin_{c \in \mathbf{C}} L_{\theta}(y|c, \mathbf{x}) + L(\theta),
\end{equation}
where each $c$ represents one possible organization of examples. $L_{\theta}(\mathbf{y}|c, \mathbf{x})$ is the codelength required to compress and transmit testing label $y$ given the organization $c$ and testing input $\mathbf{x}$. $L(\theta)$ is the codelength required to describe the model, which can be ignored during ranking since all organizations use the same model without parameter updating. 
The codelength required for data transmission can be calculated using \textit{Shannon-Huffman code}:

\begin{equation}
\label{eq:shannon}
    L_{\theta}(y|c, \mathbf{x}) = - log_2 \, p(y|c, \mathbf{x}).
\end{equation}
However, since we don't have access to testing label $y$ when ranking, the exact computation of $p(y|c, \mathbf{x})$ is impossible. To tackle this problem, we propose to compute the expectation of codelength as the surrogate:
\begin{equation}
\label{eq:expectation}
    L_{\theta}(y|c, \mathbf{x}) \approx  - \mathbb{E}_{q(y_i|Y)} log_2 \, p(y_i|c, \mathbf{x}),
\end{equation}
where $q(y_i|Y)$ is the prior of $y_i$ among all possible labels $Y$. A natural design choice of the prior is a uniform distribution, given that most datasets are label-balanced. However, since we focus on instance-level selection rather than corpus level, the likelihood $p(y_i|Y)$ can vary significantly given different samples. We thus model this term using $ p(y_i|c, \mathbf{x})$, leading to our final objective:
\begin{equation}
\label{eq:final}
    c^* = \argmin_{c \in \mathbf{C}} - \mathbb{E}_{p(y_i|c, \mathbf{x})} log_2 \, p(y_i|c, \mathbf{x}).
\end{equation}

Now that we have an interpretable metric for ranking, we can brute-force all possible permutations to obtain the optimal ranking result. Although we have significantly reduced the search space using the selection module, enumerating all organizations is still infeasible. For instance, if we want to search for the best organization that contains 8 examples, even a small candidate set of 10 examples can result in 1.8 million choices ($\textbf{A}_{10}^8$). At the current stage, we randomly sample 10 permutations for ranking. We leave it as an interesting future work to investigate how to approximate the optimal ranking better. 

\subsection{Interpretation of $ L_{\theta}(y|c, \mathbf{x})$}
Except for the compression viewpoint, we offer some other interpretations of our method here. 

\paragraph{Connection to entropy} When we use model confidence $p(y_i|c, \mathbf{x})$ as the estimation of $q(y_i|Y)$, Eq~\ref{eq:expectation} is basically calculating the entropy. Minimizing entropy is equivalent to searching for in-context examples that will lead to a skewed probability distribution. In other words, we are searching for in-context examples are will make PLMs very confident about its answer. This motivation is exactly opposite to the Local Entropy(LocalE) metric proposed by \citet{lu2022fantastically}, where they search by maximizing the entropy. 

\paragraph{Connection to cross-entropy.} Note that in this paper, we focus on instance level \icl and assume no validation set is available. However, when we have a validation set to directly compute $p(y|c, \mathbf{x})$, Eq~\ref{eq:shannon} is exactly the categorical cross-entropy loss. Hence, trying to minimize the description length of the outputs is equivalent to minimizing the usual classification loss. This reveals why compression is another viewpoint of learning.

\paragraph{Connection to mutual information.} Previous effort~\cite{blier2018description} has proved that the compression is limited by the mutual information between inputs and outputs:
$$
H(y)-\mathbb{E}_q[L(y \mid x)] \leq H(y)-H(y \mid x)=I(y ; x),
$$
where we assume the inputs and outputs follow the joint distribution $q$. Based on this finding, any successful compression of the labels is, at the same time, a direct estimation of the mutual information between input and output. This connects our method to \citet{sorensen2022information} that selects templates by maximizing mutual information. 

\paragraph{Difference to previous works.} Except for the aforementioned connections and differences, our method significantly differs from \citet{lu2022fantastically} and \citet{sorensen2022information} in that we perform instance-level selection without a validation set. Trivial extension of previous methods to our setting is impractical: \citet{lu2022fantastically} requires a validation set to compute the \textit{Global Entropy}, while the mutual information is always zero on instance-level setting according to \citet{sorensen2022information}.

\section{Experiments}
\subsection{Evaluation details}
We perform experiments across eight different NLP datasets. Unless otherwise stated, all experiments are conducted using GPT2-XL (1.5B)~\cite{radfordlanguage}. Our method is denoted as \textbf{TopK+MDL}, in which we first use TopK to retrieve 30 candidates for each sample and then randomly sample 10 organizations (each with 8 examples) for ranking using MDL. All models and datasets are loaded from HuggingFace Hub. Templates are adopted from \citet{ye2022zerogen,gao2021making} and detailed in Table~\ref{tab:templates}. We ran all experiments three times with different random seeds and reported the average accuracies. 

\paragraph{Datasets}
We consider two sentiment classification datasets~\cite{DBLP:conf/emnlp/SocherPWCMNP13}: SST-2 and SST-5, three natural language inference datasets: SNLI~\cite{bowman2015snli}, MNLI~\cite{williams2017mnli}, and QNLI~\cite{wang2018glue}, one multi-choice question answering dataset: Commonsense QA (CMS QA)~\cite{talmor2019commonsenseqa}, two topic classification datasets: TREC~\cite{trec} and AgNews~\cite{agnews}.

\paragraph{Baselines}
We compare our framework with three groups of baselines: prompting, corpus-level methods, and instance-level methods. \textbf{Prompting} is a special case of \icl without in-context examples. For corpus-level methods, we consider two methods that require a validation set: \textbf{GlobalIE}~\cite{lu2022fantastically} and \textbf{Random \& Validation}, which picks 10 random organizations for each dataset and selects the best one according to the validation performance. We also consider validation-free baselines: Mutual Information (\textbf{MI})~\cite{sorensen2022information} and a \textbf{Random} baseline that randomly initiates one organization for each dataset. 
For instance-level methods, we consider \textbf{TopK+LocalE}~\cite{lu2022fantastically}, \textbf{TopK}~\cite{liu2022makes} and a \textbf{Random} baseline that randomly selects 8 examples for each testing sample. We further add a \textbf{Majority vote} baseline that directly performs majority voting based on 8 examples retrieved by TopK.

\paragraph{Evaluation Strategy}
Due to the restricted test set access of some datasets (MNLI, QNLI, and CMS QA), we hold out a small subset (i.e., 10\%) of the training set for validation for corpus-level methods, and report results on the validation set. 
For \textsc{Prompting} and instance-level methods, we directly evaluate them on the original validation set when the test set is not available.

\subsection{Main Results}

\begin{table*}[!ht]
    \centering
    \resizebox{\textwidth}{!}{
    \begin{tabular}{l|lllllllll}
    \toprule
        \textbf{} & \textbf{SST-2} & \textbf{SST-5} & \textbf{SNLI} & \textbf{MNLI}   & \textbf{QNLI} & \textbf{Trec} & \textbf{AgNews} & \textbf{CMS QA} & \textbf{AVG} \\ \midrule
        \textbf{Prompting} & 71.38 & 29.41 & 41.23 & 39.19 & 50.44 & 13.8 & 29.75 & 39.39 &  39.32 (52.99\%$\mathbf{\uparrow}$)   \\   \midrule
         \multicolumn{10}{c}{\textbf{Corpus-level}}  \\ \midrule
         \textbf{Random} & 73.68 & 23.88 & 43.35 & 39.43 & 53.19 & 19.66 & 36.92 & 52.66 & 42.78 (40.41\%$\mathbf{\uparrow}$)   \\ 
         \textbf{Random \& Validation} & 87.86 & 40.10 & 49.27 & 43.26 & 51.12 & 32.67 & 52.01 & 53.75 & 51.25 (17.38\%$\mathbf{\uparrow}$)   \\ 
        \textbf{MI}~\cite{sorensen2022information} & 52.86 & 35.35 & 46.02 & 41.32 & 50.62 & 16.00 & 47.29 & 52.78 & 42.85 (40.63\%$\mathbf{\uparrow}$)   \\ 
        \textbf{GlobalE}~\cite{lu2022fantastically} & 87.27 & 33.21 & 46.99 & 40.46 & 57.27 & 28.53 & 52.01 & 22.42 & 49.75 (20.92\%$\mathbf{\uparrow}$)   \\ \midrule
         \multicolumn{10}{c}{\textbf{Instance-level}}  \\ \midrule
         \textbf{Random} & 77.17 & 25.65 & 43.41 & 41.17 & 53.09 & 18.33 & 32.71 & 52.93 & 43.06 (39.72\%$\mathbf{\uparrow}$)   \\
        \textbf{TopK}~\cite{liu2022makes} & 83.91 & 37.01 & 57.54 & 45.72 & 59.72 & 40.80 & \textbf{88.89} & 51.51 & 58.14 (3.48\%$\mathbf{\uparrow}$)   \\ 
        \textbf{Majority vote} & 85.34 & \underline{41.58} & 52.06 & 34.38 & 58.02 & \underline{51.60} & 60.91 & 19.57 & 50.43 (19.29\%$\mathbf{\uparrow}$)   \\ 
         \textbf{TopK+LocalE}~\cite{lu2022fantastically} & 67.12 & 31.65 & 46.78 & 41.51 & 52.66 & 36.20 & 81.88 & 53.07 & 51.36 (17.17\%$\mathbf{\uparrow}$)  \\ \hline
        \textbf{Ours} (TopK+MDL) & \textbf{91.51} & \textbf{40.27} & \textbf{58.77} & \textbf{46.56} & \textbf{61.43} & \textbf{42.47} & 87.94 & \textbf{53.15} & \textbf{60.16}  \\ 
    \bottomrule
    \end{tabular}
    }
    \caption{Evaluation results. Numbers in bold indicate the highest accuracy among all methods (except Majority vote). Numbers in the parenthesis represent the relative improvements our method achieved over baselines.} 
    \label{tab:main}
\end{table*}

From Table~\ref{tab:main}, we first observe that \icl methods outperform \textit{prompting} in most cases. However, we also note that bad in-context organizations (e.g., the random baseline) can hurt performance and make \icl performs even less well than prompting on SST-5. These results stress the importance of correct selection and permutation of in-context examples. 

We first compare our methods with corpus-level methods. As shown in Table~\ref{tab:main}, our method shows consistent and clear superiority over corpus-level baselines. This result also validates our conjecture that corpus-level methods can be sub-optimal and self-adaptive in-context examples can significantly improve \icl performance. Remarkably, our method demonstrates a 40\% relative improvement against the common practice in \icl (i.e., the Random baseline). Such improvement is encouraging as it shows that despite the surprising performance of \icl in many tasks, there is still a large room for improvement with advanced in-context example searching techniques. 

Our method still registers decent improvements on most evaluated datasets even when compared with instance-level baselines. Compared with TopK+LocalE, our method makes a 17\% relative improvement, this demonstrates the effectiveness of MDL as a ranking method. 

However, we also notice that TopK is a very competitive baseline to our method. Using semantic search to retrieve examples will result in in-context examples whose input distribution and \textit{label} are quite similar, or even identical, to the testing sample. This phenomenon leads to our hypothesis about the surprising effectiveness of TopK. First, as pointed out by \citet{xie2021explanation}, \icl can be cast as an implicit Bayesian inference process, where the PLMs implicitly infer a concept when making the prediction. Based on this theoretic finding, we deduce that semantically similar in-context examples improve prediction by providing more evidence for Bayesian inference, especially for topic classification tasks like TREC and AgNews. Second, we conjecture that providing a series of examples with the same label as the testing sample introduces a ``learning shortcut'' for PLMs and biases the results. We further examine this hypothesis below. 

\subsection{Impact of label in \icl}

\begin{figure*}
    \centering
    \begin{minipage}{0.33\linewidth}
    \subfigure[\label{fig:label}]{
    \includegraphics[width=\linewidth]{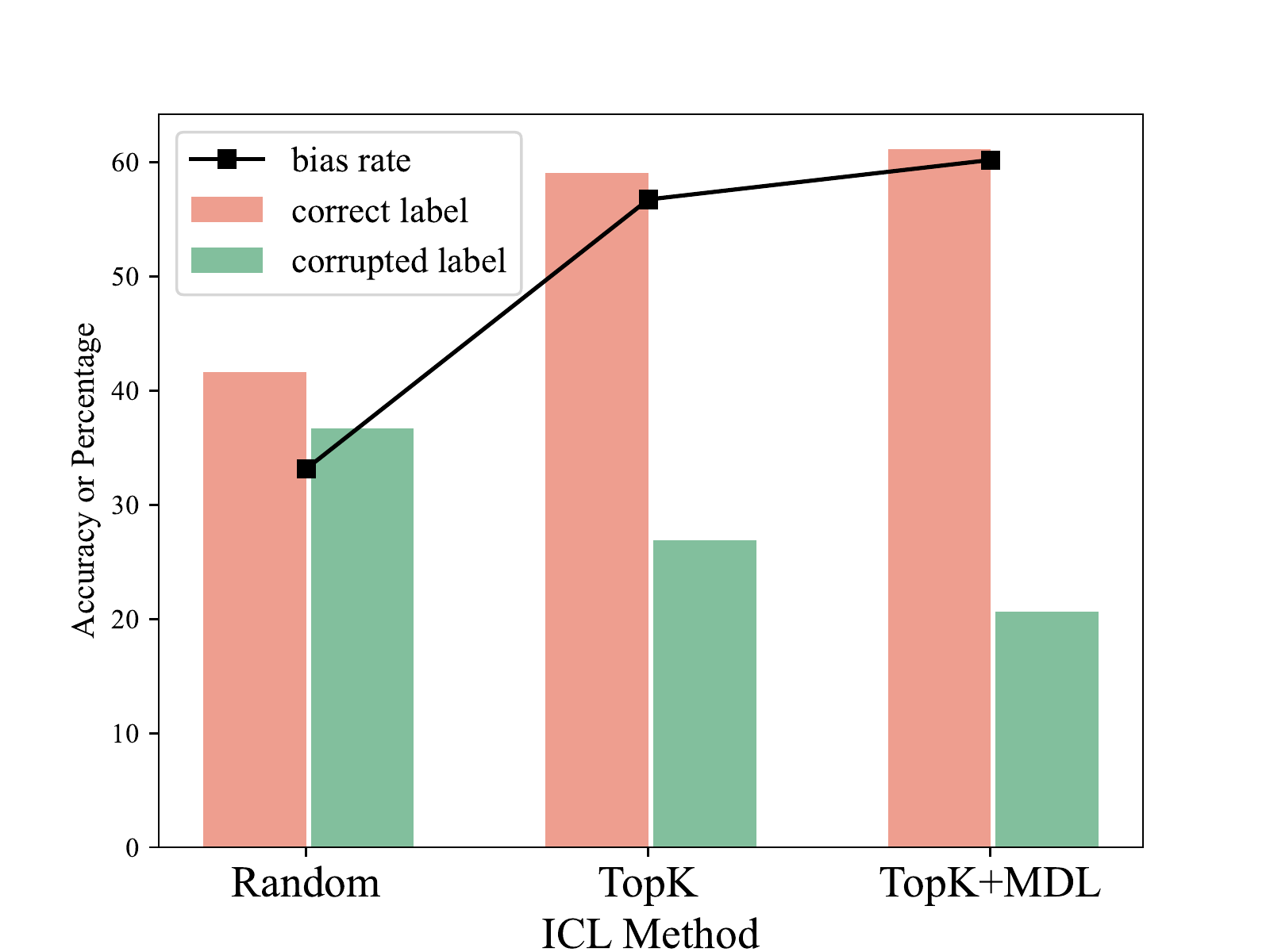}
    }
    \end{minipage}%
    \begin{minipage}{0.33\linewidth}
    \subfigure[\label{fig:few_sst2}]{
    \includegraphics[width=\linewidth]{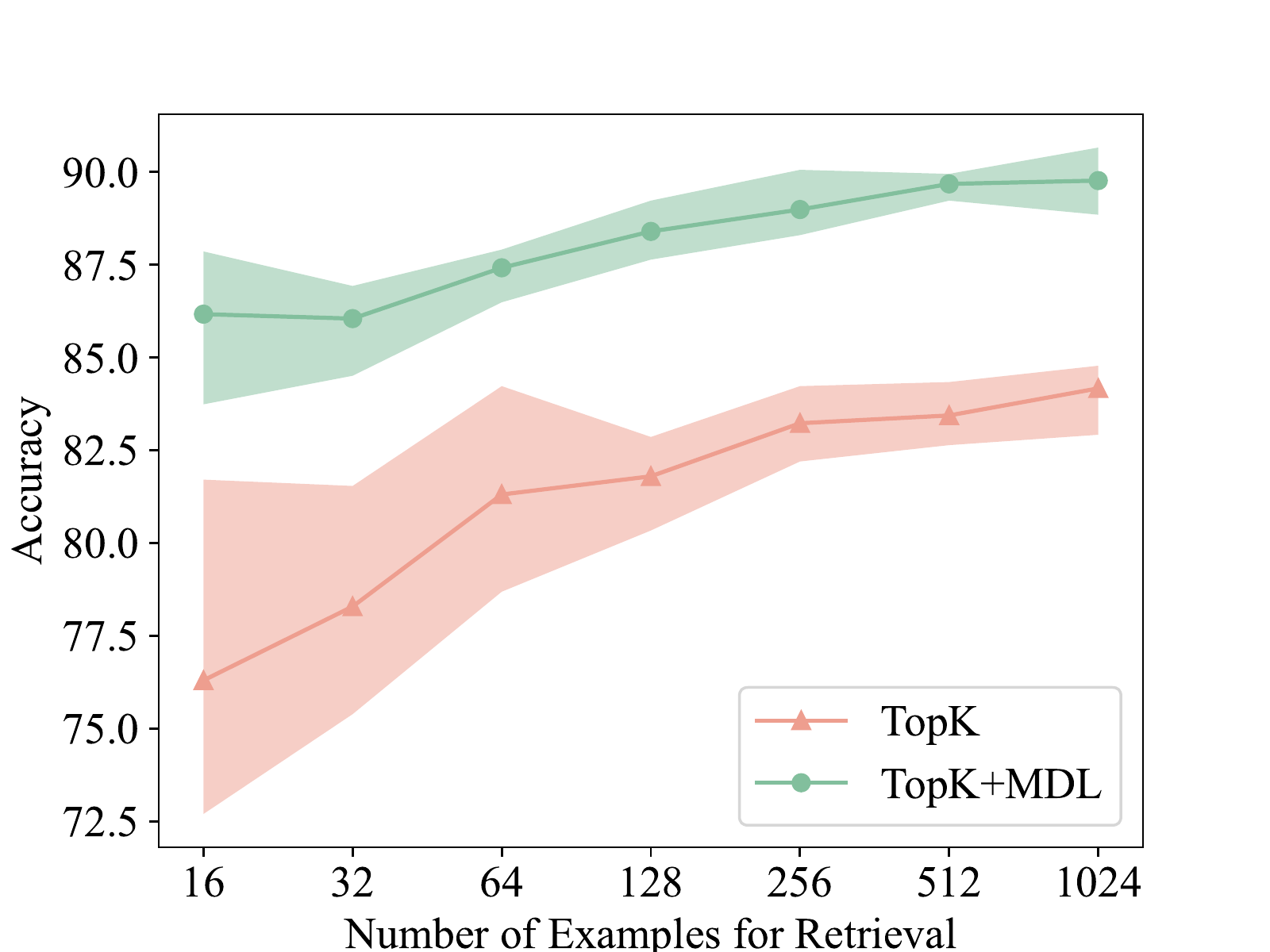}
    }
    \end{minipage}%
    \begin{minipage}{0.33\linewidth}
    \subfigure[\label{fig:few_snli}]{
    \includegraphics[width=\linewidth]{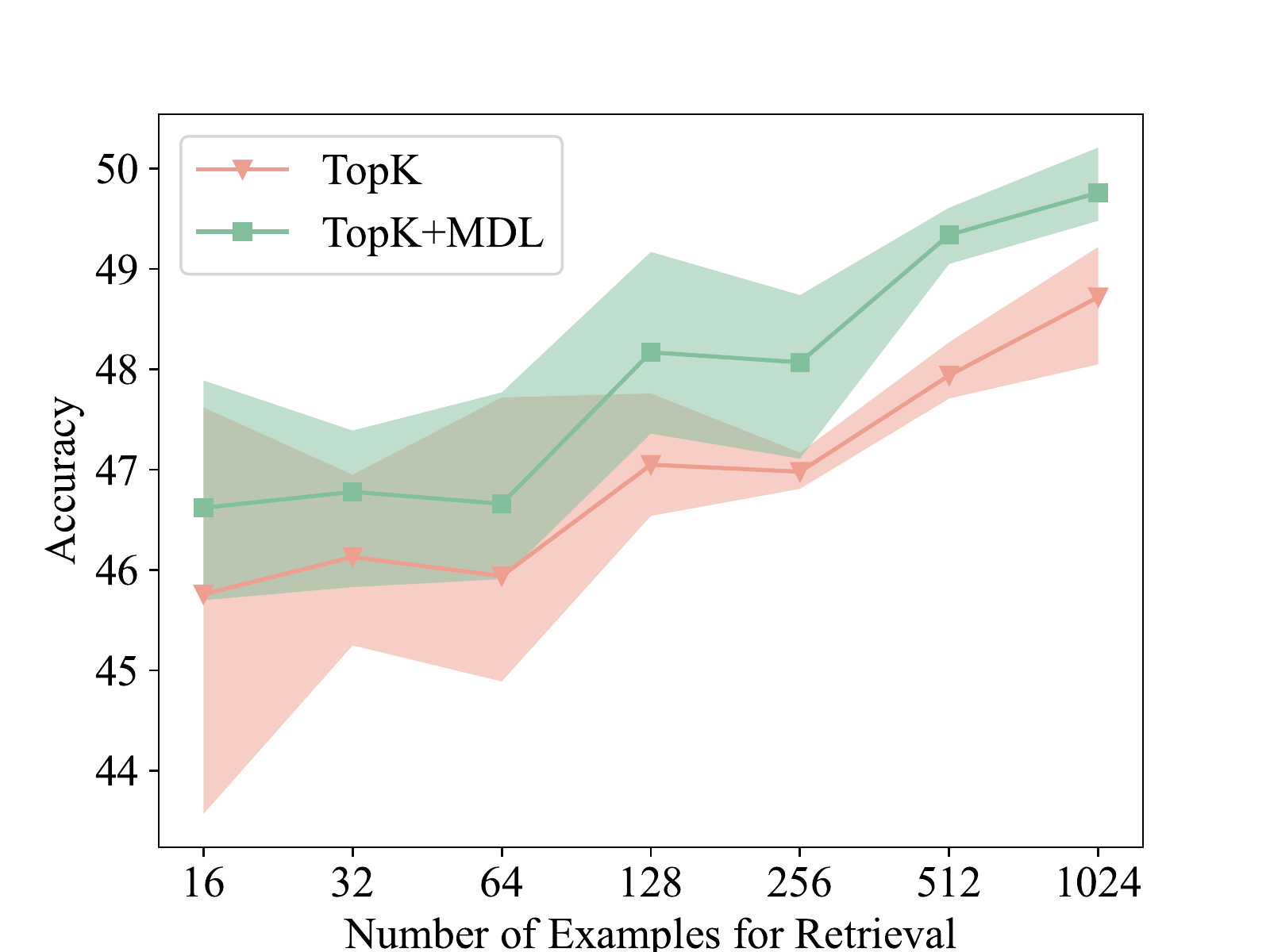}
    }
    \end{minipage}
    \caption{(a) Impact of the label in \icl. The bias rate reflects the percentage of in-context examples whose label is identical to the testing sample. (b) Few-shot results on SST2. (c) Few-shot results on SNLI.}
    \label{fig:analysis}
\end{figure*}

To investigate the impact labels have on \icl, we calculate \textit{bias rate}. Given a testing sample $(\textbf{x},y)$ and its in-context examples, the bias rate represents the percentage of in-context examples whose label is identical to $y$.  As shown in Figure~\ref{fig:label}, the bias rate positively correlates with the performance. We conduct a more fine-grained exploration by corrupting the label space and breaking the input-label alignment. We corrupt the labels by exchanging label words between classes, e.g., exchanging label words between positive and negative classes in sentiment classification. As in Figure~\ref{fig:label}, we observe a clear performance drop with corrupted labels, which negatively correlates with the bias rate.  These results suggest that in-context examples' labels could significantly impact \icl performance. Recent debates~\cite{min2022rethinking,kim2022ground} on the effect of label distribution focus on corpus-level \icl, and our findings complement their studies.

\section{Analysis}
The observed benefits of our method raise the natural question of why and how it helps and whether the same performance improvements can be transferred to other PLMs or prompts.
In this section, we conduct comprehensive experiments and analyses to understand the strength and weaknesses of our method.

\subsection{When a large set of annotated examples is not available}
\label{sec:few-shot}

Despite the surprising performance of ICL, a large-scale training set is not always available for retrieval in practice.
To address this concern, we conduct experiments under the few-shot setting. We randomly sample 16, 32, 64, 128, 256, 512, and 1024 examples as the candidates for searching. We select two representative tasks (SST2 and SNLI) for evaluation and run each experiment three times with different random seeds. 

As shown in Figure~\ref{fig:few_sst2} and \ref{fig:few_snli}, our method consistently outperforms the strong baseline TopK as in the full-data setting.   This demonstrated the general applicability of our method in both full-data and few-shot scenarios. We also observe that the performance steadily increases with the growing number of annotated examples.

\begin{figure*}[ht]
    \centering
    \begin{minipage}{0.33\linewidth}
    \subfigure[\label{fig:aba_a}]{
    \includegraphics[width=\linewidth]{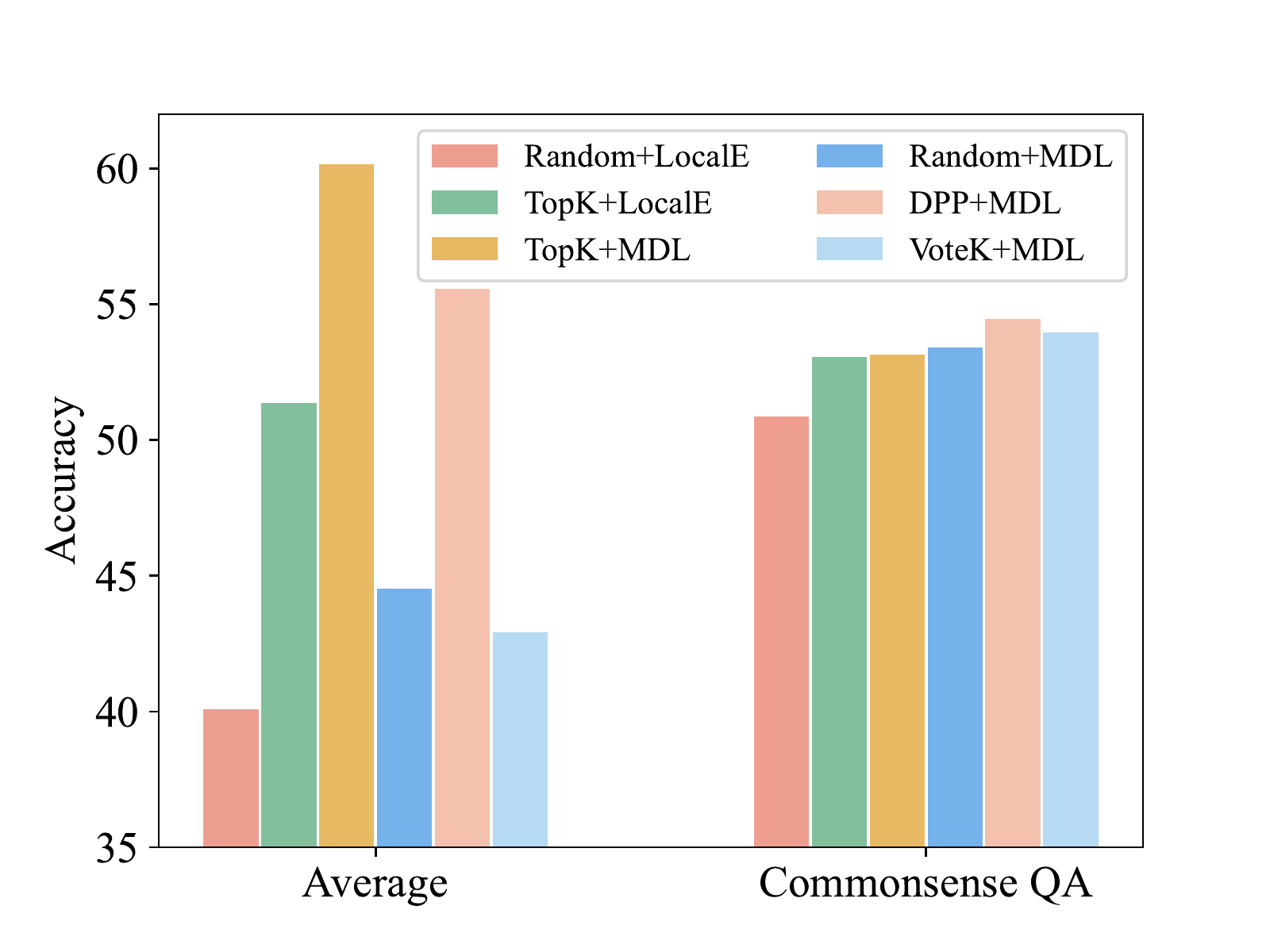}
    }
    \end{minipage}%
    \begin{minipage}{0.33\linewidth}
    \subfigure[\label{fig:aba_b}]{
    \includegraphics[width=\linewidth]{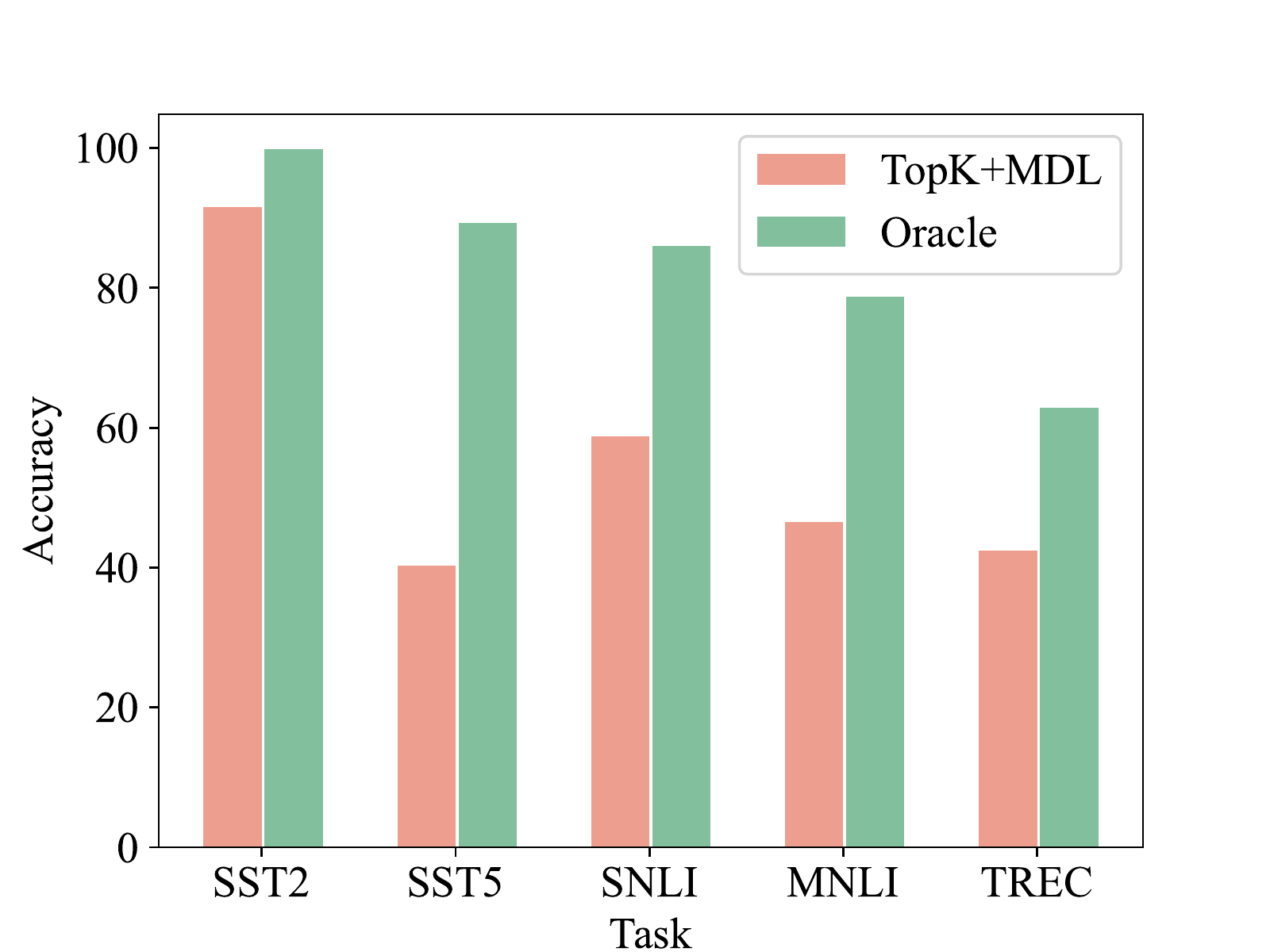}
    }
    \end{minipage}%
    \begin{minipage}{0.33\linewidth}
    \subfigure[\label{fig:aba_c}]{
    \includegraphics[width=\linewidth]{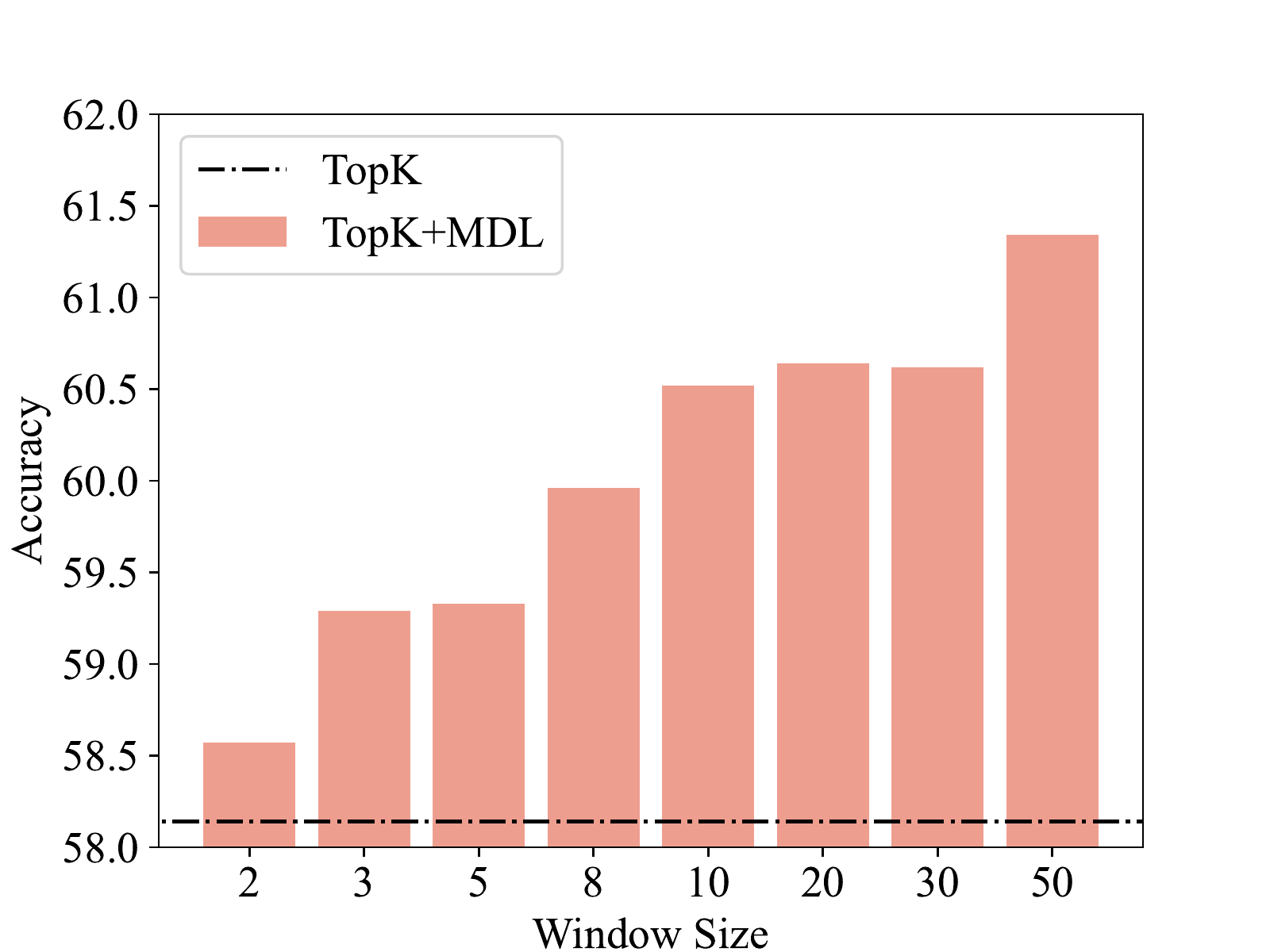}
    }
    \end{minipage}
    \begin{minipage}{0.33\linewidth}
    \subfigure[\label{fig:aba_d}]{
    \includegraphics[width=\linewidth]{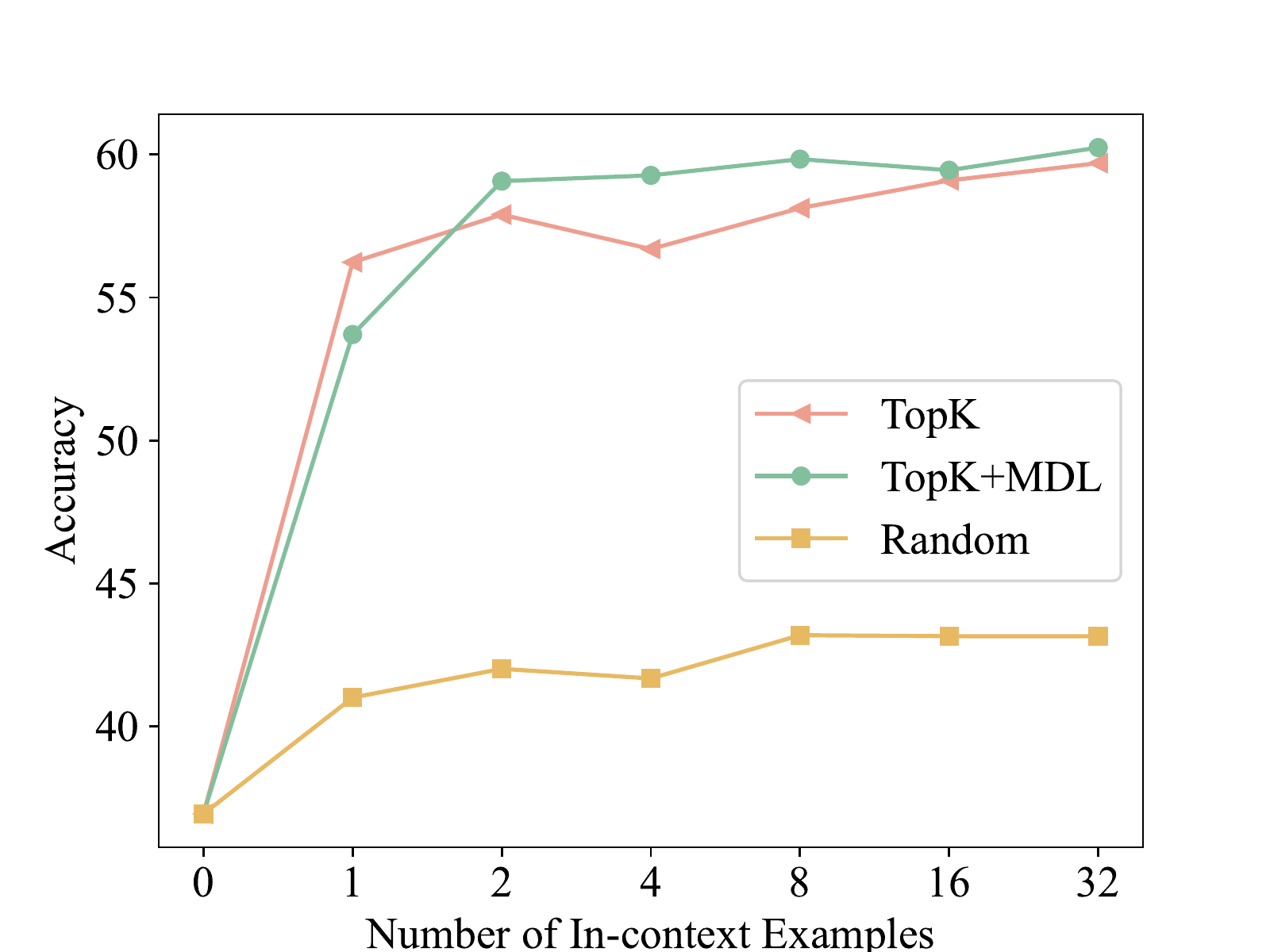}
    }
    \end{minipage}%
    \begin{minipage}{0.33\linewidth}
    \subfigure[\label{fig:aba_e}]{
    \includegraphics[width=\linewidth]{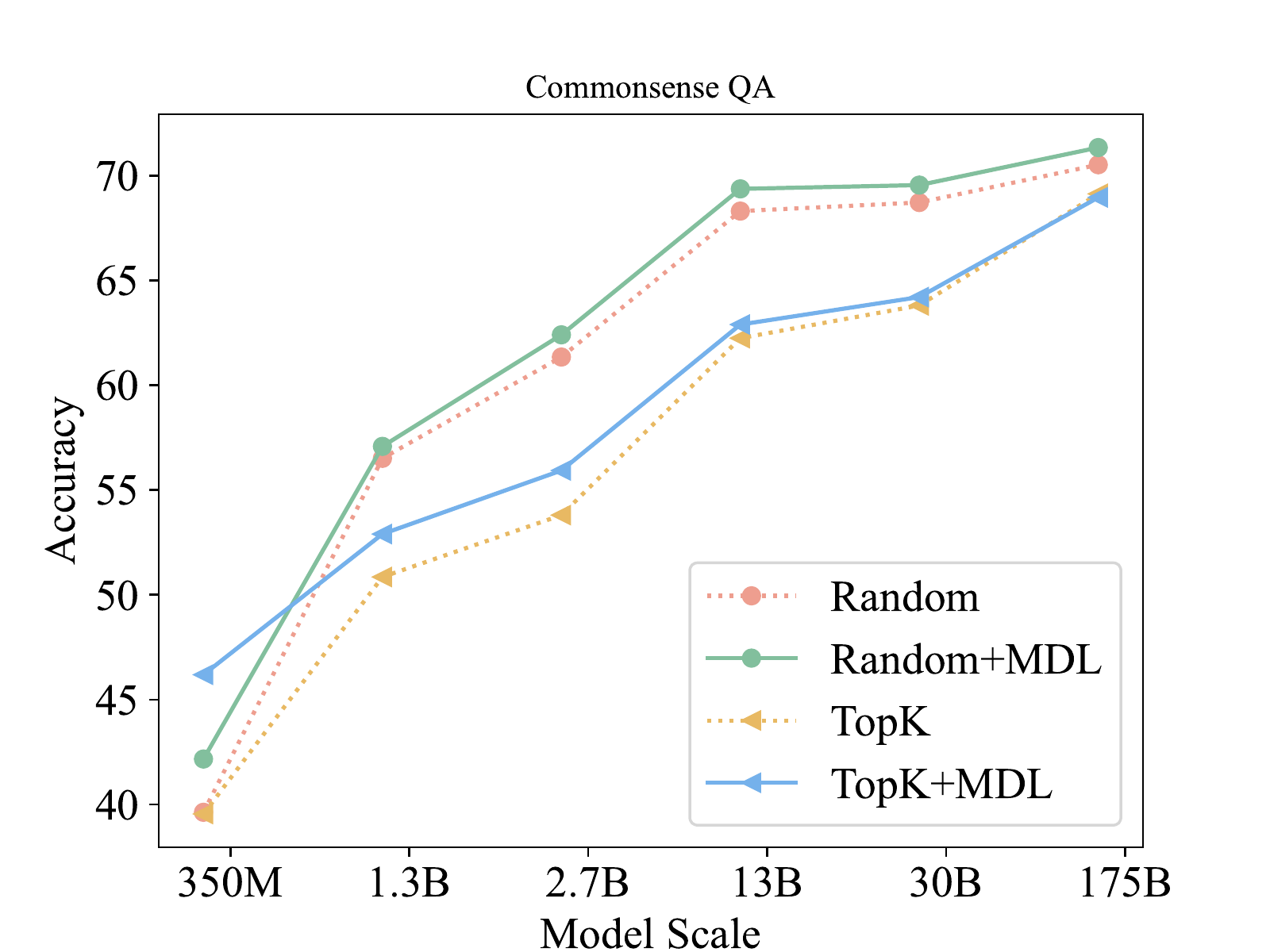}
    }
    \end{minipage}%
    \begin{minipage}{0.33\linewidth}
    \subfigure[\label{fig:aba_f}]{
    \includegraphics[width=\linewidth]{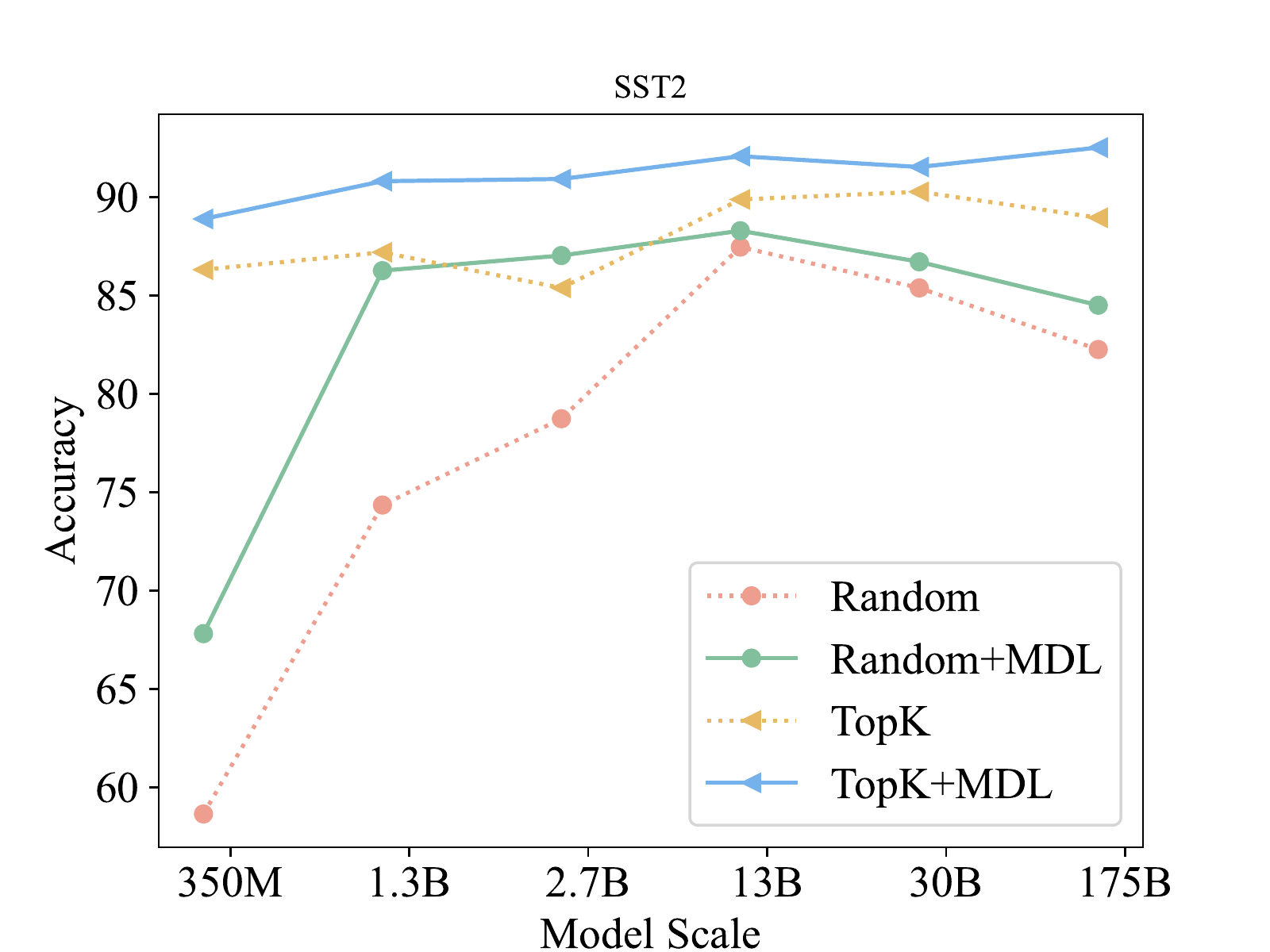}
    }
    \end{minipage}%
    \caption{(a) impact of different selection methods. (b) the accuracy of our ranking method. (c) impact of window size(number of permutations to be ranked). (d) impact of the number of in-context examples. (e,f) impact of model scales on Commonsense QA and SST2.}
    \label{fig:analysis}
\end{figure*}

\subsection{Impact of selection methods}
\label{sec:selection}
We conduct most experiments using the popular TopK method for candidate example selection. Here we evaluate three other alternatives: random, DPP and VoteK. Figure~\ref{fig:aba_a} shows that using TopK for example selection outperforms all other alternatives on average.  However, we also observe that the superiority of TopK is mainly in simple classification tasks with limited label space.  On multi-choice tasks like Commonsense QA, all three alternatives outperform TopK (right side of Figure~\ref{fig:aba_a}). Note that although multi-choice tasks are also classification tasks, they have a huge label space like NLG tasks. The frustration of TopK on multi-choice tasks suggests that  the popular TopK method does not work well for tasks with large label space and searching for better selection methods holds immense prospects, and therefore remains an interesting field of further research.

\subsection{Accuracy of ranking method}
\label{sec:oracle}
In our ranking module, we randomly select 10 different organizations for each testing sample and use MDL to select the best-performing one in an unsupervised manner. 
Despite the superior performance of MDL, the accuracy of using MDL for in-context example ranking has not been discussed. To understand the ranking accuracy of MDL, we assume a perfect ranking method \textit{oracle}, which can always select the organization that leads to correct prediction if there is any. In the implementation, we first obtain predictions for all 10 organizations. If at least one prediction matches the ground truth, we consider this testing example solvable by \textit{oracle}. As shown in Figure~\ref{fig:aba_b}, there are significant performance gaps between oracle and TopK+MDL. Although such oracle performance only exists theoretically, it's still encouraging to see the enormous promise of \icl: with better selection and ranking methods (e.g., supervised methods~\cite{rubin2022learning}), we might be able to bridge the performance gap between \icl and finetuning. 

\begin{table}[]
    \centering
    \resizebox{\linewidth}{!}{
    \begin{tabular}{l|cccc}
    \toprule
    \textbf{Dataset} & \textbf{TopK} & \textbf{TopK+MDL} & \textbf{TopK+LocalE} & \textbf{Random} \\ \midrule
     \textbf{SST-2}  &  0.6861(83.91) &	0.6810(91.51) & 0.6928(67.12) &	0.6918(77.17) \\
     \textbf{SNLI}   &  1.0981(57.54) &	1.0929(58.77) &	1.0983(46.78) &	1.0974(43.41)  \\
     \textbf{CMS QA} & 4.9883(51.51)  &	4.9371(53.15) &	4.9692(53.07) &	4.9629(52.93) \\
     \textbf{Trec}   & 5.5618(40.80)  &	5.4496(42.47) &	5.7434(36.20) &	5.7859(18.33) \\
    \bottomrule
    \end{tabular}
    }
    \caption{Average MDL of each method.}
    \label{tab:mdl}
\end{table}

We investigate the correlation between MDL and accuracy by selecting four representative datasets and reporting the MDL of each method. As shown in Table~\ref{tab:mdl}, a smaller MDL generally indicates a higher accuracy (in the brackets). This validates the effectiveness of MDL as the criterion for in-context example searching. It's also interesting to see that tasks with lower MDL are generally easier to learn (as explained in \S~\ref{sec:ranking}), thus ICL has a better performance. 

\subsection{Impact of hyperparameter}
\label{sec:hyper}
In this subsection, we investigate how different hyperparameters affect our performance. 

\paragraph{Increasing the window size of our method can steadily boost performance, by trading efficiency for better performance.} We vary window size (i.e., number of organizations to be ranked per sample) from 2 to 50, and report the average accuracy. As shown in Figure~\ref{fig:aba_c}, the performance steadily increases with the window size. We even observe gains when the window size is two. In particular, on tasks with short input lengths like SST2, using a window size of 2 already shows a clear gain (+3.19 in accuracy) over TopK. However, the improvement is achieved by sacrificing efficiency, i.e., window size hits 50 means performing forward passing for the test set 50 times. Together with findings above, we conclude that we must keep improving the accuracy of ranking methods to achieve a better efficiency-effectiveness trade-off. 

\paragraph{Increasing the number of in-context examples boosts accuracy for most tasks.} We gradually increase the number of in-context examples (denoted as $N$) from 0 (prompting) to 32. From Figure~\ref{fig:aba_d}, we see that increasing $N$ consistently improves the performance on average. We also note that the random baseline reaches the performance plateau from $N=8$. Such contradictions suggest that when analyzing the impact of $N$, the organization of examples is critical. Sometimes 
 we find increasing $N$ not helpful because we are not using the ``right'' organization. Our results raise an interesting question for future research: can we achieve finetuning-level performance by using thousands or even more examples as context?  

\paragraph{Larger model size does not guarantee better performance, but our method can bring consistent improvements over strong baselines.} We use OPT and vary the model size from 350M to 175B. We have a mixed observation that blindly applying huge models does not always result in the best performance. For simple tasks like SST2 (see Figure~\ref{fig:aba_f}), we reach the performance plateau after 1.3B. And for SNLI, a 30B OPT even outperforms the 175B counterpart. Large models are powerful when dealing with complex tasks like Commonsense QA. From Figure~\ref{fig:aba_e}, we can see steady and significant improvement whenever we scale up the model size. In addition, our method brings consistent improvements over baselines regardless of model sizes on all tasks evaluated. 

\subsection{Robustness}
\begin{figure}
    \centering
    \includegraphics[width=0.8\linewidth]{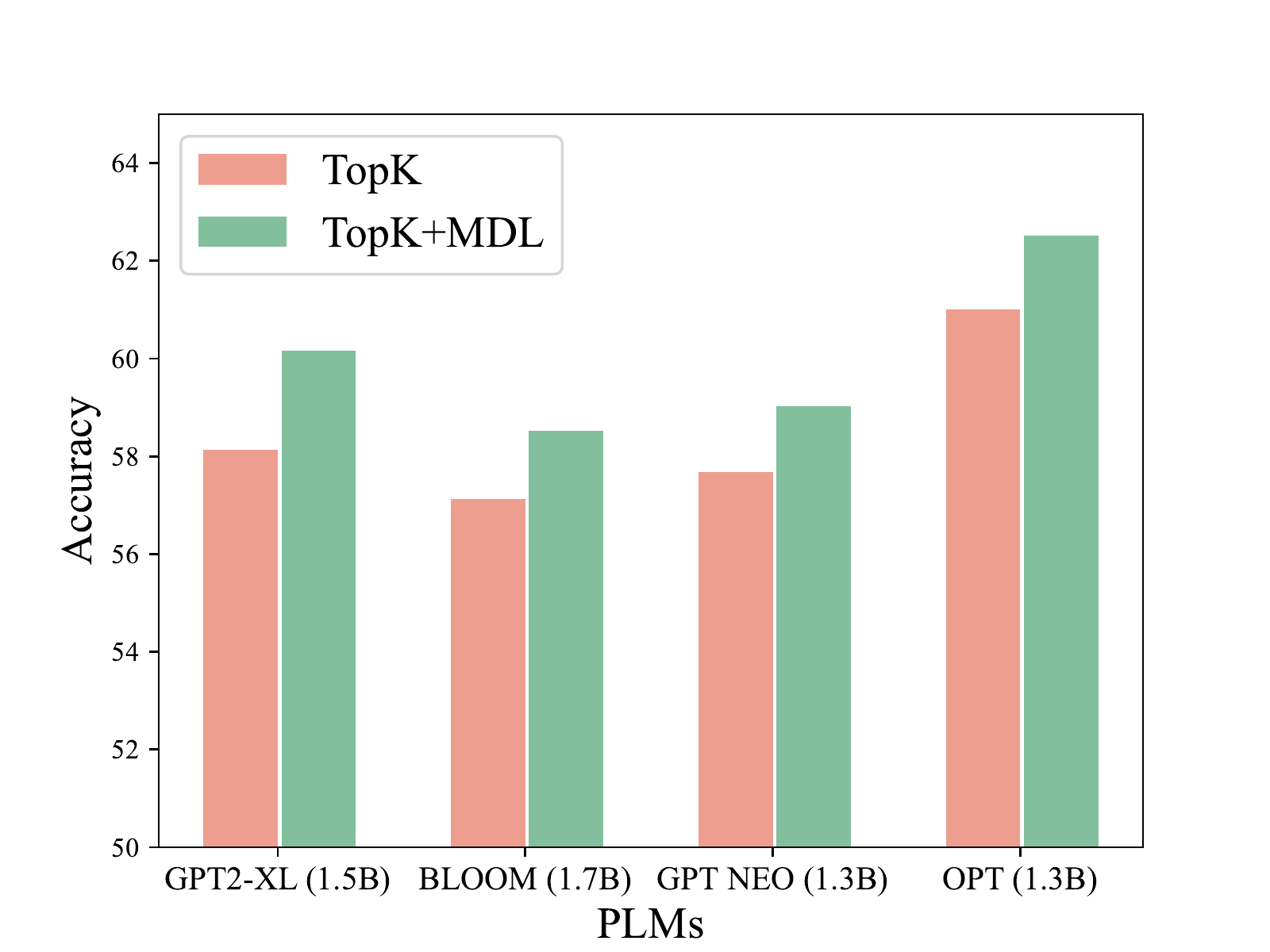}
    \caption{The average performance of TopK and our method on different PLMs.}
    \label{fig:plm}
\end{figure}
\paragraph{Generability across different PLMs.} We explore how our method generalizes between different PLMs. We average our results across datasets and present the results in Figure~\ref{fig:plm}. On four different PLMs tested, our method consistently and significantly outperforms the strong TopK baseline. Overall, we have observed that our method is robust across various datasets and PLMs. 

\paragraph{Generability across different prompts.} As sensitivity to prompt engineering is a key weakness of \icl, we evaluate the robustness of our method given different templates. We select two representative tasks (i.e., SST2 and SNLI) to conduct experiments, each with three different templates. As shown in Figure~\ref{fig:prompts}, our method is robust given different prompting templates. But still, the differences in prompting templates cause large variances in performance. The findings here motivate a line of research that simultaneously searches for the best template and in-context organization, which is rarely explored in the literature.


\section{Conclusion}
This paper proposes a new paradigm for \icl: self-adaptive ICL. Unlike existing efforts that universally use one single example organization on all testing samples, we propose a general two-stage select-then-rank framework to search in-context examples at the instance-level. We instantiate this framework with an information-theory-driven ranking algorithm. Empirical results suggest that self-adaptive in-context learning can significantly outperform the common practice in \icl by a large margin. We reveal the great potential of self-adaptive in-context learning and point out several interesting research problems in method analysis.

\section{Limitation}
Despite the demonstrated effectiveness of self-adaptive ICL, this new paradigm suffers from the following limitations. (I) As we discussed in \S~\ref{sec:hyper}, due to the large search space, we need to trade efficiency for effectiveness. So how to balance the efficiency-effectiveness 
trade-off is an important decision choice to make when deploying self-adaptive ICL methods. (II) As shown in \S~\ref{sec:few-shot}, the gains of our method shrink when the size of the retrieval set gets smaller. To maximize performance, we require a high-quality retrieval set, which might not always be available when dealing with unseen tasks in practice. We also note that both limitations can be alleviated with better selection and ranking algorithms. 

The remarkable performance of our method should partially attribute to the powerful TopK selection method, so we also discuss the limitation of TopK here. Despite its popularity, our analysis (\S~\ref{sec:selection}) reveals that TopK's effectiveness is limited to simple NLU tasks with limited label space, and it does not work well with tasks with large or even infinite label space (QA, multi-choice, and NLG).  This limitation signals a new direction for ICL research: we need better selection methods to adapt ICL methods to more tasks.

\section{Acknowledgement}
Yaoxiang, Zhiyong, and Jiacheng participate in coding and discussion. Yaoxiang and Zhiyong conduct the evaluation and analysis. Zhiyong leads the project and writes this manuscript. We want to thank members of Shark-NLP and reviewers for their valuable feedback. This work is partially supported by the National Key R\&D Program of China(NO.
2022ZD0160100), and in part by Shanghai Committee of Science and Technology (Grant No. 21DZ1100100).

\bibliography{anthology,custom}
\bibliographystyle{acl_natbib}

\appendix

\section{Datasets}
Dataset information is detailed in Table~\ref{tab:dataset}.

\begin{table}[]
    \centering
    \resizebox{\linewidth}{!}{
    \begin{tabular}{l|ccc}
    \toprule
    \textbf{Dataset} & \textbf{Task} & \textbf{Data Split} \\ \midrule
     \textbf{SST-2}  &  Sentiment Classification & 6920/872/1821/ \\
     \textbf{SST-5}  &  Sentiment Classification & 8544/1101/2210 \\
     \textbf{SNLI}   &  Natural Language Inference & 550152/10000/10000 \\
     \textbf{MNLI}   &  Natural Language Inference & 392702/19647/19643 \\
      \textbf{QNLI}  &  Natural Language Inference & 104743/5463/5463 \\
     \textbf{Trec}   &  Topic Classification & 5452/0/500 \\
     \textbf{AgNews} &  Topic Classification & 120000/0/7600 \\
    \textbf{CMS QA}  &  Commonsense Question Answering & 9471/1221/1140 \\ 
    \bottomrule
    \end{tabular}
    }
    \caption{Details of datasets.}
    \label{tab:dataset}
\end{table}


\section{Impact of hyperparameters}

\begin{figure}
    \centering
    \includegraphics[width=0.8\linewidth]{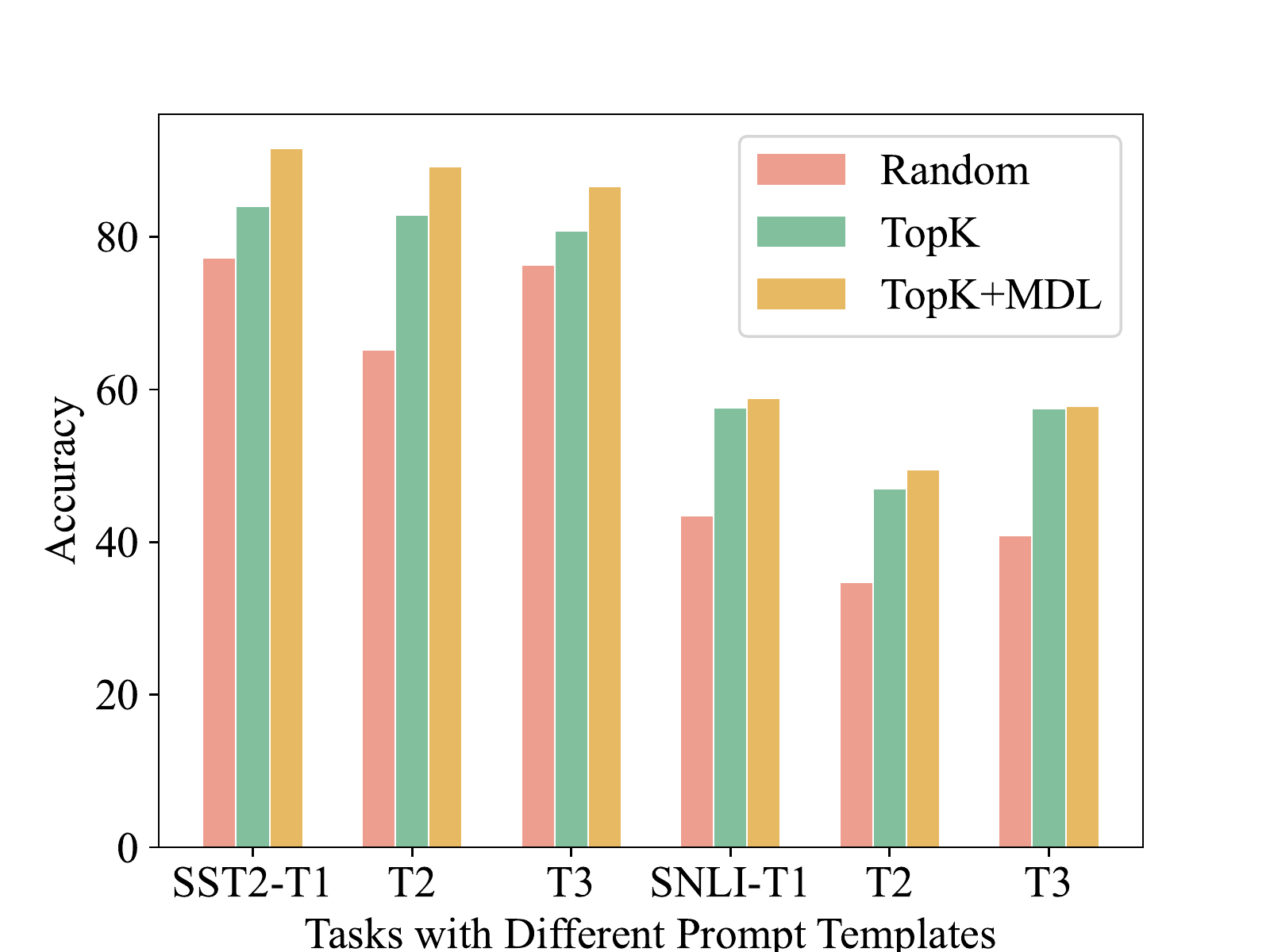}
    \caption{Results of TopK and our method on SST2 and SNLI, using different prompts.}
    \label{fig:prompts}
\end{figure}

\begin{figure}[ht]
    \centering
    \resizebox{0.9\linewidth}{!}{
    \includegraphics{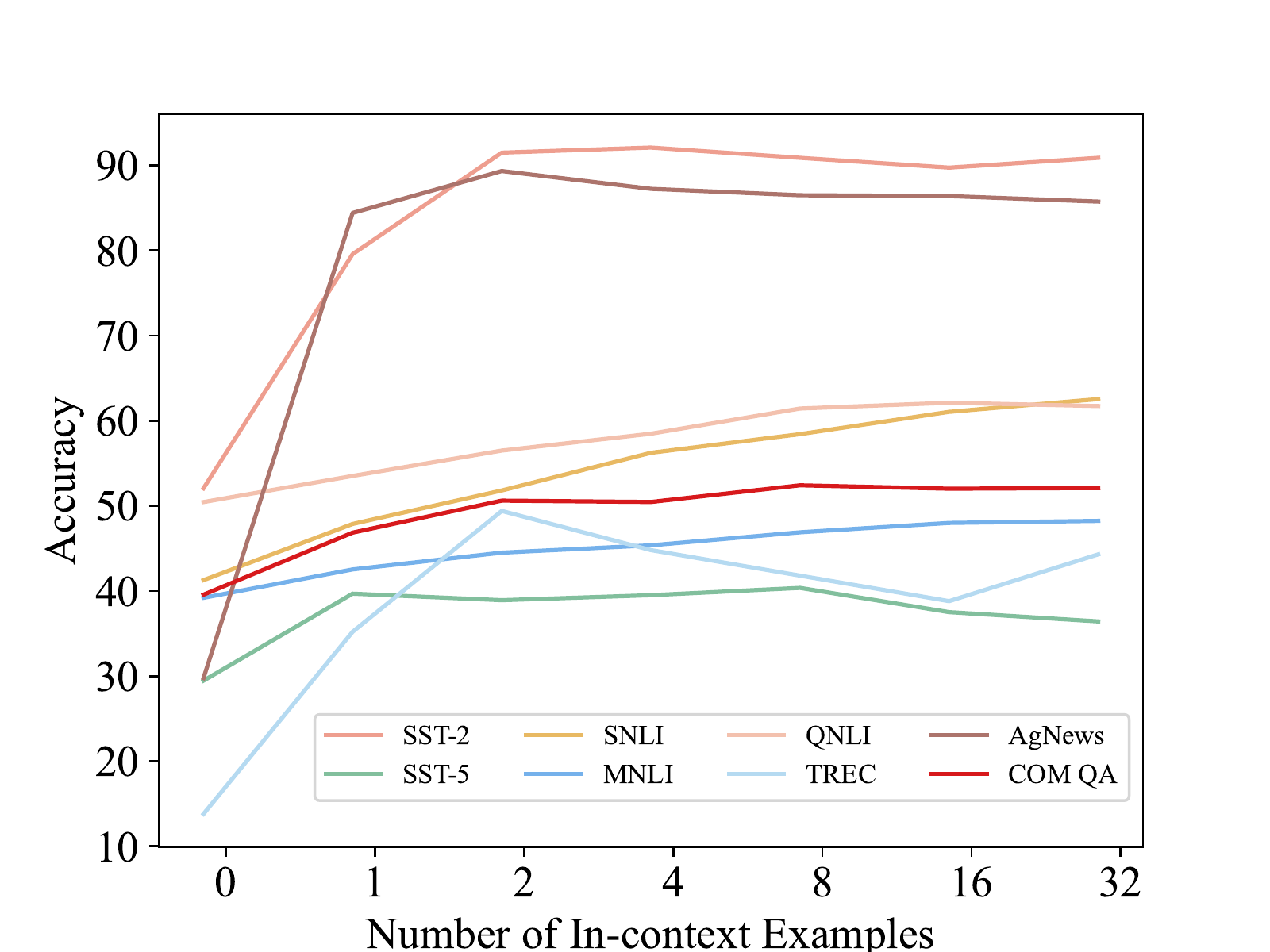}
    }
    \caption{Impact of number of in-context examples. }
    \label{fig:nicl_full}
\end{figure}

\begin{figure}[ht]
    \centering
    \resizebox{0.9\linewidth}{!}{
    \includegraphics{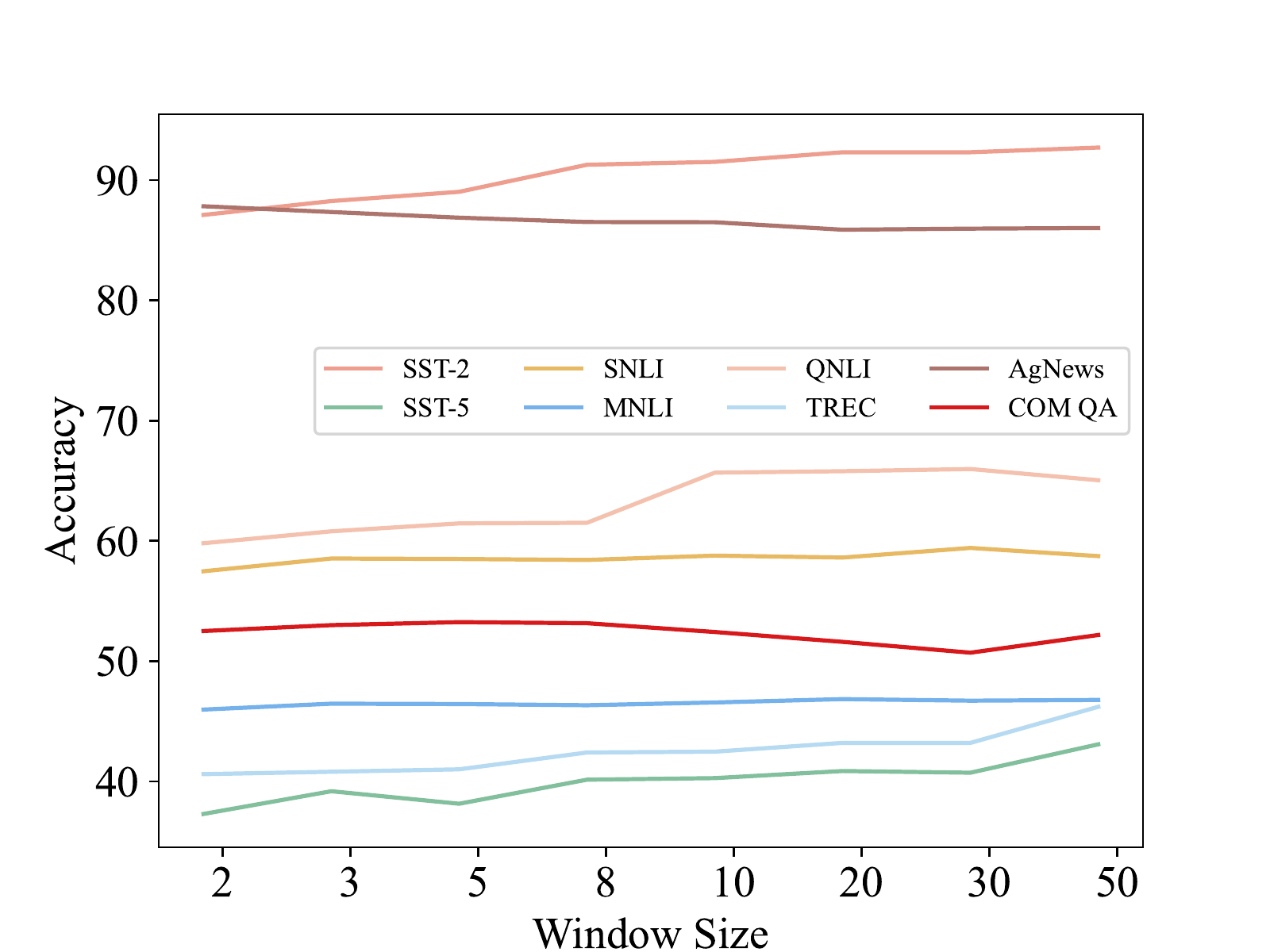}
    }
    \caption{Evaluation results with different window sizes (number of permutations to be ranked).}
    \label{fig:win_full}
\end{figure}

The results of adjusting the number of in-context examples and window size are shown in Figure~\ref{fig:nicl_full} and \ref{fig:win_full}, respectively. 

\section{Templates}
\begin{table*}[t ]
\centering
\resizebox{0.8\linewidth}{!}{
\begin{tabular}{lll}
\toprule
\textbf{Task} & \textbf{Prompt} & \textbf{Class} \\
\hline
\multirow{2}{*}{SST-2} & Positive Movie Review: "<X>"  & Positive \\ 
& Negative Movie Review: "<X>"  & Negative \\
\midrule
\multirow{5}{*}{SST-5}
& "<X>" It is terrible.  & Very Negative \\ 
& "<X>" It is bad.  & Negative \\
& "<X>" It is OK.  & Neutral \\ 
& "<X>" It is good.  & Positive \\
& "<X>" It is great.  & Very Positive \\ 
\midrule
\multirow{3}{*}{SNLI\, \& \, MNLI}
& <X1>? Yes, <X2>  & Entailment \\ 
& <X1>? Maybe, <X2>  & Neutral \\
& <X1>? No, <X2>  & Contradiction \\ 
\midrule

\multirow{2}{*}{QNLI}
& <C> Can we know <X>? Yes.  & Entailment \\ 
& <C> Can we know <X>? No.  & Contradiction \\ 
\midrule
\multirow{6}{*}{TREC}
& "<X>" It is about abbreviation.  & ABBR \\ 
& "<X>" It is about entity.  & ENTY \\ 
& "<X>" It is about description and abstract concept.  & DESC \\ 
& "<X>" It is about human being.  & HUM \\ 
& "<X>" It is about location.  & LOC \\ 
& "<X>" It is about numeric value.  & NUM \\ 

\midrule

\multirow{4}{*}{AgNews}
& "<X>" It is about world.  & World \\ 
& "<X>" It is about sports.  & Sports \\ 
& "<X>" It is about business.  & Business \\ 
& "<X>" It is about science and technology.  & Sci/Tech \\ 
\midrule
\multirow{5}{*}{Commonsense QA}
& Answer the following question:\,<X>\,  Answer: <A>.  & A \\ 
& Answer the following question:\,<X>\,  Answer: <B>.  & B \\ 
& Answer the following question:\,<X>\,  Answer: <C>.  & C \\ 
& Answer the following question:\,<X>\,  Answer: <D>.  & D \\ 
& Answer the following question:\,<X>\,  Answer: <E>.  & E \\ 
\midrule

\end{tabular}
}
\caption{Templates of tasks. Placeholders (e.g., <X> and <A>) will be replaced by real inputs or answers (in Commonsense QA).}
\label{tab:templates}
\end{table*}

The templates used in this paper are detailed in Table~\ref{tab:templates}.

\end{document}